\documentclass[letterpaper, 10 pt, journal, twoside]{IEEEtran}
\ifCLASSINFOpdf
  \usepackage[pdftex]{graphicx}
\else
\fi
\usepackage{amsmath}
\usepackage{amssymb,amsfonts}

\usepackage{optidef}  

\usepackage{algorithmic}

\usepackage{float}  

\usepackage{multirow}
\usepackage{textcomp}
\usepackage{xcolor}
\makeatletter
\let\NAT@parse\undefined
\makeatother
\usepackage[numbers,sort]{natbib}

\newcommand{\crossmat}[1]{\begin{bmatrix}\,#1\,\end{bmatrix}_\times}

\newcommand{\ppartial}[2]{\frac{\partial{#1}}{\partial{#2}}}
\DeclareMathOperator{\trace}{Tr}

\begin{document}
%
\title{Object-Centric Task and Motion Planning in Dynamic Environments}
%
%
%

\author{Toki Migimatsu and Jeannette Bohg
\thanks{The authors are with the Department of Computer Science, Stanford University, Stanford, CA 94309 USA (e-mail: takatoki@cs.stanford.edu).}
}
\maketitle

\begin{abstract}
We address the problem of applying \textit{Task and Motion Planning} (TAMP) in real world environments. TAMP combines symbolic and geometric reasoning to produce sequential manipulation plans, typically specified as joint-space trajectories, which are valid only as long as the environment is static and perception and control are highly accurate. In case of any changes in the environment, slow re-planning is required. We propose a TAMP algorithm that optimizes over Cartesian frames defined relative to target objects. The resulting plan then remains valid even if the objects are moving and can be executed by reactive controllers that adapt to these changes in real time. We apply our TAMP framework to a torque-controlled robot in a pick and place setting and demonstrate its ability to adapt to changing environments, inaccurate perception, and imprecise control, both in simulation and the real world.  
\end{abstract}


%
\IEEEpeerreviewmaketitle

\section{Introduction}
%
%
%
%
\IEEEPARstart{R}{obot} manipulation tasks often require a combination of symbolic and geometric reasoning. For example, if a robot wants to grasp an object outside of its workspace, a symbolic planner might devise a plan to use a long stick to bring the object closer, while a motion planner would determine where to grasp the stick and how to use it to push the object. {\em Task and Motion Planning\/} (TAMP) algorithms address this challenge of interleaving symbolic and geometric reasoning. They often produce full trajectories for the robot to execute. However, due to the PSPACE-hard complexity of motion planning \cite{reif1979complexity}, generating these trajectories can take on the order of minutes, and if the environment changes, these trajectories may become invalid and require replanning.

\begin{figure}
    \centering
    \includegraphics[width=\columnwidth]{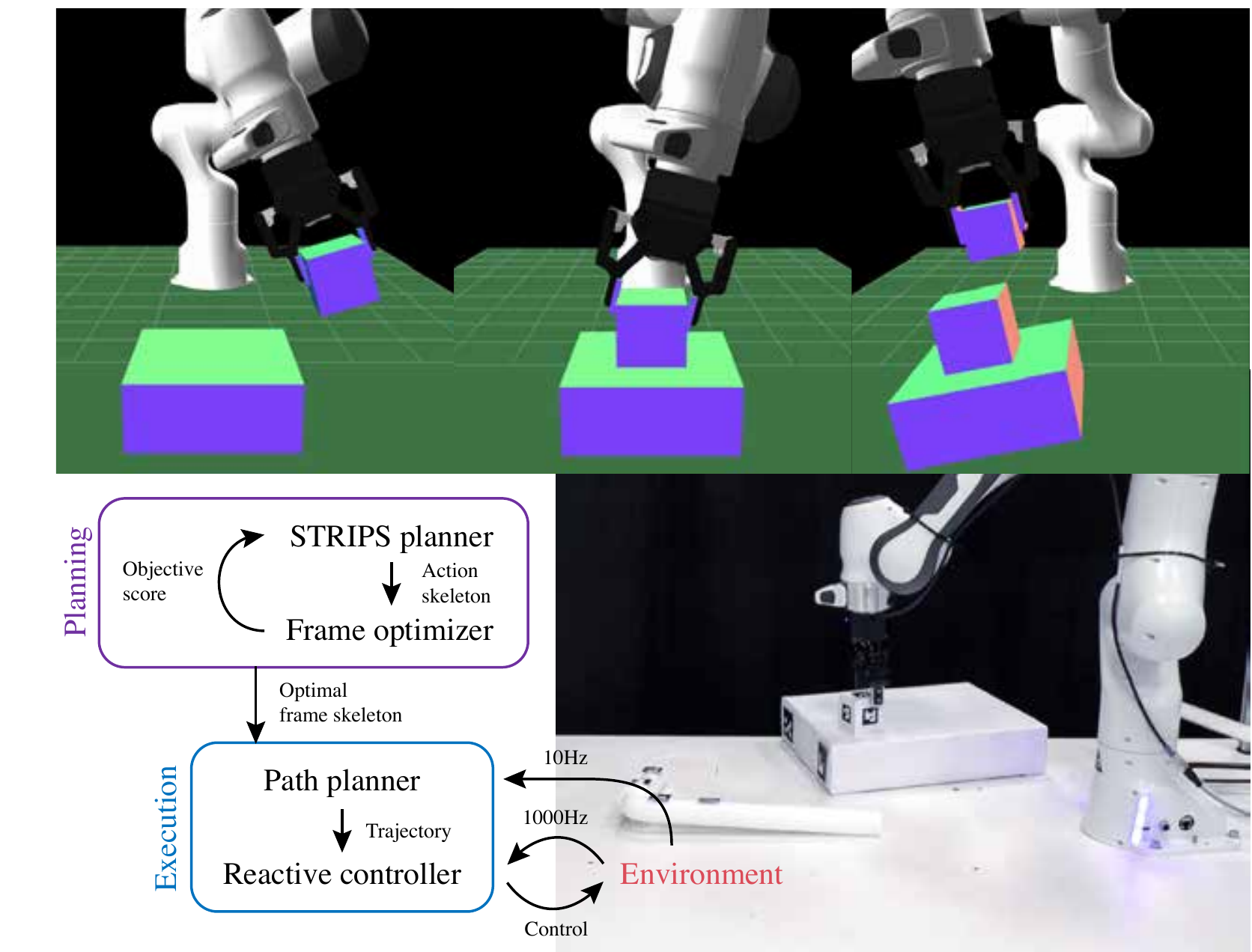}
    \caption{The proposed TAMP algorithm plans over relative Cartesian frames for integration with reactive controllers that can adapt to changes in the environment in real-time. The top row demonstrates the robot's ability to complete the block stacking task without stopping to re-plan even when the poses of the grasped object and target location change. The bottom row summarizes our TAMP framework and shows an example execution on the real robot (see supplementary material for a video).}
    \label{fig:header}
\end{figure}

In this paper, we combine TAMP with reactive control to enable robot manipulation in real-world scenarios where the environment may change during execution, perception may be inaccurate, and robot control may be imprecise. Instead of generating full joint-space trajectories like most TAMP algorithms, our approach returns desired object poses at key timepoints. These object poses are defined relative to target frames, which may be moving. Optimizing over relative poses instead of joint configurations at this stage of planning facilitates integration with local controllers that can react to changes in the environment in real time. For example, if the robot needs to pick up an object but specifies this pick action using a desired joint configuration, this joint configuration becomes invalid as soon as the object's position changes, leading to unintended contact or external disturbances. However, if the planner outputs a desired pose of the end-effector relative to the object, then this relative pose is still valid no matter where the object moves, eliminating the need to stop and replan. Figure~\ref{fig:header} provides an overview of the proposed framework.

In addition to facilitating integration with reactive controllers, planning over object poses also has the benefit of being a more natural representation for object manipulation. The geometric constraints for actions such as \textit{pick}, \textit{place}, and \textit{push} are straightforward in Cartesian space, which simplifies the process of defining new manipulation actions.

The main theoretical contributions of this work are twofold. First, we introduce a formulation of TAMP using object-centric frames that work with reactive controllers to accommodate for changing environments, inaccurate perception, and imprecise control. Second, we derive the \textbf{SE}(3) functions necessary for optimizing object-centric frames in pick and place applications. We then apply this TAMP framework to a manipulation setting where a robot can \textit{pick}, \textit{place}, and \textit{push}, and demonstrate it working on a real robot where the environment state is visually tracked with RGB-D cameras.

\section{Related Work}
Task and motion planning has been widely studied since the introduction of STRIPS, the first automated task planner \cite{fikes1971strips}. One common approach of TAMP algorithms is to interleave the symbolic and geometric search processes by calling a motion planner at every step of the symbolic search and tentatively assigning geometric parameters to the current symbolic state before advancing to the next one \cite{plaku2010sampling, cambon2009hybrid}.

One issue that arises from interleaving symbolic and geometric search is when a state found in the search process is symbolically legal but geometrically infeasible due to a previous instantiation of geometric parameters. \citet{lagriffoul2012constraint,lagriffoul2014efficiently} introduce the concept of geometric backtracking, where the planner must search for alternative geometric instances until a solution is found. They propose an algorithm that maintains interval bounds over placement parameters to reduce the search space of geometric backtracking. \citet{bidot2017geometric} limit geometric backtracking with heuristics based on statistics for kinematic violations and collisions with movable objects. \citet{de2013towards} use {\em Hierarchical Task Networks\/} (HTNs) to perform symbolic search using hierarchically abstracted tasks, performing HTN backtracking whenever geometric backtracking fails. {\em Hierarchical Planning in the Now\/} (HPN) \cite{kaelbling2010hierarchical} similarly uses a hierarchical approach, but interleaves planning with execution, such that primitive actions at the lowest level of the hierarchy are executed as soon as they are reached by the search algorithm. This requires that the actions themselves be reversible when backtracking is necessary.

While integrating geometric search with symbolic search can prune out large sections of the symbolic space, calling motion planning with every symbolic search increment can be costly, and even disadvantageous if most states are geometrically feasible \cite{lagriffoul2013combining}. An alternative approach is to perform geometric search only on full candidate symbolic plans, or action skeletons. One method of conveying geometric feasibility back to the symbolic search is to construct predicates from the geometric parameters \cite{lozano2014constraint,ferrer2017combined,srivastava2014combined}.
\citet{dantam2016incremental} use an incremental satisfiable modulo theory (SMT) solver to incrementally generate action skeletons and invoke a motion planner in between for validation. Similarly, \citet{zhang2016co} formulate the TAMP problem as a {\em Traveling Salesman Problem\/} (TSP), using a motion planner in the inner loop of a symbolic planner to update the weights of the TSP.

While most TAMP algorithms use sampling-based motion planners, Toussaint et al. \cite{toussaint2015logic,toussaint2018differentiable} use nonlinear optimization to identify geometric parameters for given action skeletons. This approach, called {\em Logic Geometric Programming\/} (LGP), is what we extend in this work to handle dynamic environments. Advantages of using nonlinear optimization include the ability to consider all geometric parameters simultaneously without the need for backtracking, as well as the ability to encode complex goals in the objective such as minimizing the energy of the robot or maximizing the height of a stack of blocks.

A common limitation of the TAMP methods above is that planning is slow, taking on the order of tens of seconds to minutes. To this end, \citet{garrett2018sampling} reduce the sampling space of motion planning by conditionally sampling on factoring patterns in the action skeletons. \citet{wells2019learning} use a {\em Support Vector Machine\/} (SVM) to estimate the feasibility of actions to guide the symbolic search, only calling the motion planner on action skeletons classified by the SVM as feasible.

Even with these improvements in efficiency, the theoretical complexity of TAMP makes it impractical to execute inside a real-time closed-loop controller. Therefore, common assumptions in TAMP are that objects are static, object poses are accurately perceived before planning, and robot commands are executed precisely. Otherwise, errors can accumulate, becoming problematic when a robot needs to grasp an object that it imprecisely placed earlier in the plan, for example. To deal with perception errors, \citet{suarez2018interleaving} incorporate a symbolic action where the robot examines an object up close when perception uncertainty is high. However, this approach is still unable to handle dynamic environments.

We address these limitations by planning over relative object poses, so that an open-loop plan generated by the TAMP algorithm can still be executed by a real-time closed-loop controller even if object poses change during execution.

\section{LGP Background}

\subsection{General LGP Formulation}

An LGP is comprised of two subproblems: a STRIPS problem~\cite{fikes1971strips} specified using first-order logic that operates in a discrete domain, and a nonlinear trajectory optimization problem that operates in a continuous domain. The goal of an LGP is first to find a sequence of $K$ discrete actions $a_{1:K}$ that results in a discrete state trajectory $s_{1:K}$ such that the final state $s_K$ satisfies a set of goal propositions $\mathfrak{g}$. Second, the LGP needs to find a sequence of continuous control inputs $u(t)$ across time $t = [0, T]$ such that the resulting state trajectory $x(t)$ satisfies the requirements of the discrete states $s_{1:K}$. This can be framed as the following optimization problem: 
\begin{argmini}[1]
    {a_{1:K}, u(t)}{h(x(T)) + \int_0^T g(x(t), u(t)) \,dt}{}{}
    \addConstraint{x(0)}{= x_{init}, \quad s_K \vDash \mathfrak{g}}
    \addConstraint{\dot{x}(t) }{= f_{path}(x(t), u(t), s_{k(t)})}{\quad t \in [0, T]}
    \addConstraint{\dot{x}(t_k)}{= f_{switch}(x(t_k), u(t_k), a_{k})}{\quad k = 1, \dots, K}
    \addConstraint{s_k}{\in \operatorname{succ}(s_{k-1}, a_k)}{\quad k = 1, \dots, K}
\end{argmini}

$h$ is the terminal cost and $g$ is the trajectory cost typical in optimal control problems. $f_{path}$ describes the continuous system dynamics, which change depending on the discrete state $s_{k(t)}$ at a given timestep $t$. An example of this dependency is when a robot manipulator throws a ball---while the manipulator is holding the ball, the ball's trajectory is determined by the manipulator's dynamics, but after release, it follows unconstrained projectile dynamics. $f_{switch}$ describes the instantaneous system dynamics at timesteps $t_k$ when the discrete action $a_k$ changes. An example of such a constraint is when a robot uses a stick to hit a ball---at the time of impact, the ball's acceleration is determined by the instantaneous impulse imparted by the stick.

In STRIPS planning, each action $a_k$ defines a set of preconditions that must be true before the action is performed and a set of postconditions that will be true after the action is performed. These discrete transition dynamics are encoded in the $\operatorname{succ}$ constraint using first-order logic.

While LGPs can be cast as Mixed-Integer Nonlinear Programs, such problems are in general undecidable and cannot be solved exactly~\cite{jeroslow1973there}. \citet{toussaint2015logic} proposes an algorithm that approximately solves LGPs by breaking it into a multi-part process. First, a tree search is performed in the STRIPS domain to find a candidate action skeleton $a_{1:K}$. Next, the action skeleton spawns a trajectory optimization problem whose constraints are defined by the action skeleton and corresponding symbolic states. This trajectory optimization is first solved over key timesteps $t_k$ for $k = 1, \dots, K$, and then subsequently over all timesteps $t = [0, T]$. This optimization will either produce a full trajectory $x_{1:T}$ with an objective score, or fail to find a solution, which means the candidate action skeleton is not physically feasible. The tree search continues until the user decides to terminate, at which point the trajectory with the smallest objective score gets returned.


\subsection{Joint Space Formulation}

\citet{toussaint2018differentiable} formulate the trajectory optimization subproblem as a $k$-order Markov Optimization (KOMO) problem, which discretizes time and represents time derivatives of position (up to $k$th order) with their discrete equivalents. The state space and control inputs are defined to be the manipulator configuration $q$ extended with an extra 6-dof free body joint for each object the robot is manipulating. This augmented configuration is represented by $\bar{q}_t \in \mathbb{R}^{n + 6 m_t}$, where $m_t$ is the number of objects the robot is manipulating at timestep $t$. Note that the size of $\bar{q}_t$ changes with $t$. 
\begin{argmini}
    {\bar{q}_{0:T}}{h(\bar{q}_T) + \sum_{t=0}^T g(\bar{q}_{t-2:t})}{}{}
    \addConstraint{\bar{q}_{-2:0}}{= \bar{q}_{init}}
    \addConstraint{f_{path_{k(t)}}(\bar{q}_{-2:t})}{= 0}{\quad t = 1, \dots, T}
    \addConstraint{f_{switch_k}(\bar{q}_{-2:t_k})}{= 0}{\quad k = 1, \dots, K}
\end{argmini}

The objective term $g(\bar{q}_{t-2:t})$ is a function that can take into account the position, velocity, and acceleration at each timestep. The sum of these functions across all timesteps results in a banded Hessian that can be inverted efficiently for Gauss-Newton methods. However, the constraint functions depend on the entire trajectory of joint positions---for example, the constraint associated with placing an object on a table is affected by how it was picked up at a previous timestep, since this determines the configuration of the object in the robot's end-effector. This coupling between timesteps makes the Jacobians for these constraints complex, since the constraint functions are dependent on all previous timesteps of the trajectory, not just the timestep of the constraint itself.

\section{Cartesian Formulation}
In this section, we introduce the new Cartesian frame formulation that enables adaptation to dynamic environments.
At the most basic level, a manipulation task can be reduced to controlling a point relative to some reference frame, e.g., controlling the tip of a hammer relative to the head of a nail. Instead of optimizing over augmented joint configurations of the robot and manipulated objects, the optimization variables in this new formulation are defined to be the relative pose between the control and target frames at each timestep. In a real system, this pose could be observed from a combination of 3D pose estimation and the forward kinematics of the robot.

Let $\operatorname{control}(t)$ and $\operatorname{target}(t)$ be the control and target frames determined by the action at time $t$ in the action skeleton. The relative pose of $\operatorname{control}(t)$ in $\operatorname{target}(t)$ is represented by a 6-dof variable $\xi_t = \begin{pmatrix} \xi_{t_p} & \xi_{t_r} \end{pmatrix}^T$, where $\xi_{t_p} \in \mathbb{R}^3$ is the position and $\xi_{t_r} \in \mathbb{R}^3$ is the orientation in axis-angle form. The axis-angle representation is chosen for the orientation because it does not have any constraints over its parameters and does not suffer from kinematic singularities like gimbal lock. Although it has representation singularities at rotations of $k\pi$, this should not pose a problem for the optimizer. Note that $\xi_t$ is a representation of the Euclidian lie algebra $\mathfrak{se}(3)$.
\begin{argmini}
    {\xi_{0:T}}{h(\xi_{0:T}) + \sum_{t=1}^T g_t(\xi_{0:t})}{}{}
    \addConstraint{\xi_0}{= \xi_{init}}
    \addConstraint{f_{path_{k(t)}}(\xi_t)}{= 0}{\quad t = 1, \dots, T}
    \addConstraint{f_{switch_k}(\xi_{t_k})}{= 0}{\quad k = 1, \dots, K}
\end{argmini}

In this Cartesian formulation of the problem, we only consider the positions of the frames in the trajectory and ignore velocities and accelerations. Dynamic motions that require controlling velocities and accelerations are more likely to require high control frequencies that this stage of the trajectory planning cannot offer. 
Therefore, to enable real-time reactive behavior, the trajectory planner considers only key positions in the trajectory (e.g., at all the switch timesteps) and leaves the higher derivatives up to controllers designed for each action.

\begin{figure}
    \centering
    \includegraphics[width=\columnwidth]{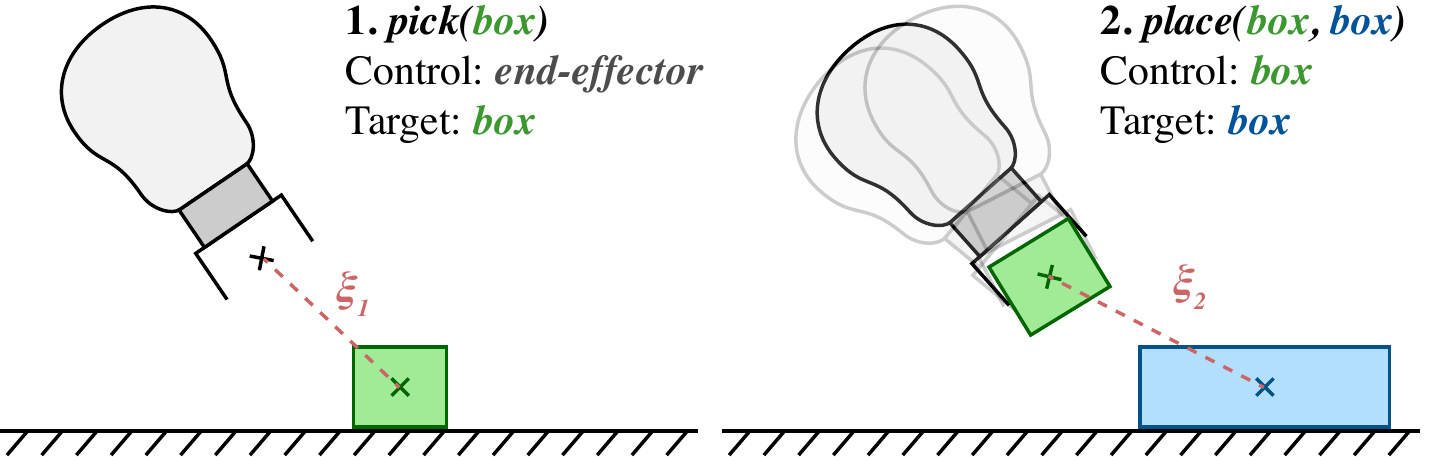}
    \caption{Optimization variables $\xi_t$ define the 6-dof pose of the control frame relative to the target frame. The place task is unaffected by the grasp executed by the preceding pick task, since it only involves the relative pose between the two boxes. This decouples the action constraints in the optimization problem.}
    \label{fig:decoupling}
\end{figure}

One advantage of optimizing over local frames, in addition to facilitating integration with local controllers and enabling reactive behavior to changing object positions, is that the optimization constraints become decoupled in time. For example, with joint configuration variables, the constraint for placing an object on a table depends on the configuration used to pick up the object, since this determines the pose of the object relative to the end-effector. However, in the Cartesian formulation, if we define the variable for the place constraint to be the relative pose between the object and the table, then this constraint no longer depends on the configuration used to pick up the object, since the relative pose of the object in the end-effector is irrelevant (see Figure~\ref{fig:decoupling}). This decoupling simplifies the Jacobians for the constraints, since each constraint is only a function of a configuration variable at a single time step.

Instead, the cross-time dependency gets pushed onto the objective function, which may measure something like the total distance travelled by the end-effector, described below.

\subsection{Objective Function}
The objective function for our trajectory optimization problems measures a combination of the linear and angular distance travelled by the end-effector. Let $\xi \in \mathbb{R}^{6(T+1)}$ be the vector concatenation of $\xi_0, \dots, \xi_T$. Given scaling factors $\alpha, \beta \in \mathbb{R}$, the combined distance between timesteps $t-1$ and $t$ is
\begin{align}
\begin{split}
    g_t(\xi)
        &= \alpha \left\| x_{ee}(\xi; t) - x_{ee}(\xi; t-1) \right\|_2^2 \\
        &\quad+ \beta \left\| \log\left( \phi_{ee}^{-1}(\xi; t-1) \,\phi_{ee}(\xi; t) \right) \right\|_2^2
\end{split}
\end{align}
where $x_{ee}(\xi; t)$ and $\phi_{ee}(\xi; t)$ compute the absolute position and orientation, respectively, of the end-effector at time $t$ given the entire history of relative pose variables $\xi$. $\log$ is the \textbf{SO}(3) logarithmic map that in this case computes the shortest axis-angle rotation between $\phi_{ee}(\xi; t-1)$ and $\phi_{ee}(\xi; t)$.

Such an objective function requires mapping the relative variables $\xi_t$ to the absolute pose of the end-effector in the world frame, which may depend on the control actions leading up to the current timestep. This makes the gradient for the objective function more complex, but this may be outweighed by the benefit of simpler constraint Jacobians, which greatly simplifies the process of defining new actions for the optimization framework. The mapping from relative poses to absolute poses is described in detail below.

\subsection{Absolute Pose}

\begin{figure}
    \centering
    \includegraphics[width=\columnwidth]{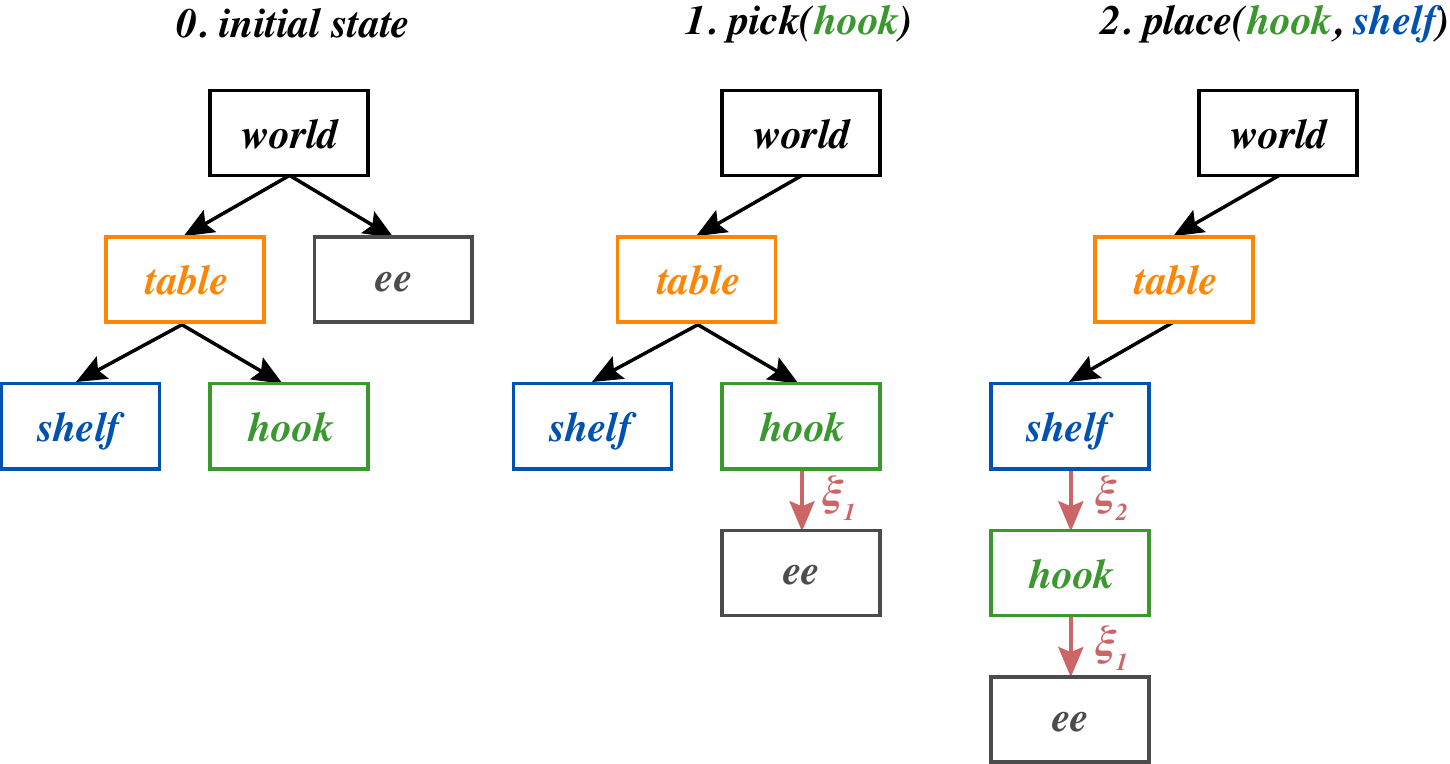}
    \caption{Diagram of the kinematic tree. In the initial state, the hook and shelf are on the table, and the table and end-effector are positioned with respect to the world. During $pick(hook)$, $\xi_1$ defines the end-effector's pose relative to the hook such that the hook becomes the end-effector's parent. During $place(hook, shelf)$, $\xi_2$ defines the hook's pose relative to the shelf, but the end-effector is still fixed to the hook via $\xi_1$. Thus, computing the absolute pose of the end-effector at $t = 2$ involves composing the relative poses down the chain from the end-effector to the world.}
    \label{fig:kinematic_tree}
\end{figure}

The variable $\xi_t$ defines the pose of $\operatorname{control}(t)$ relative to $\operatorname{target}(t)$. This means that $\operatorname{control}(t)$ is kinematically attached to $\operatorname{target}(t)$, because any pose change of $\operatorname{target}(t)$ will result in the same pose change of $\operatorname{control}(t)$. This means that computing the absolute world pose of $\operatorname{control}(t)$ requires computing the pose of $\operatorname{target}(t)$ first. The pose of $\operatorname{target}(t)$ may be defined relative to its own kinematic parent, forming a tree structure with the world frame at the root.

The kinematic tree changes at each timestep, when a manipulation action changes the kinematic parent of the control frame to a new target frame (see Figure~\ref{fig:kinematic_tree}). This change lasts for all subsequent timesteps until the same control frame is manipulated again. For example, an action that places a ball on the table at time $t$ defines $\operatorname{control}(t)$ to be the ball frame and $\operatorname{target}(t)$ the table frame. Until the ball is manipulated by another action, such as $pick$, the ball will remain fixed to the table, with its relative pose defined by $\xi_t$.

At any given timestep $t$, the pose of frame $i$ relative to its parent $\lambda(i;t)$ is either equivalent to its relative pose at the previous timestep or is defined by a new pose $\xi_t$. Formally, this can be written with the recursive definition
\begin{subequations} \label{eq:x_xi_R_xi}
\begin{align}
    {}^{\lambda(i)} x_i(\xi;t)
        &= \begin{cases}
            \xi_{t_p} & i = \operatorname{control}(t) \\
            {}^{\lambda(i)}x_i(\xi; t-1) & \text{otherwise}
        \end{cases} \label{eq:x_xi} \\
    {}^{\lambda(i)} R_i(\xi; t)
        &= \begin{cases}
            \exp(\xi_{t_r}) & i = \operatorname{control}(t) \\
            {}^{\lambda(i)}R_i(\xi; t-1) & \text{otherwise}
        \end{cases} \label{eq:R_xi}
\end{align}
\end{subequations}
where ${}^{\lambda(i;t)}x_i : \mathbb{R}^{6(T+1)} \mapsto \mathbb{R}^{3}$ and ${}^{\lambda(i;t)}R_i : \mathbb{R}^{6(T+1)} \mapsto \mathbb{R}^{3 \times 3}$ are the mappings from relative poses $\xi$ to the position and orientation, respectively, of frame $i$ in its parent $\lambda(i;t)$. $\exp$ is the \textbf{SO}(3) exponential map that converts the axis-angle representation $\xi_r$ to its rotation matrix representation.

We can compute the pose of frame $i$ in any frame $j$ by applying \eqref{eq:x_xi_R_xi} through the kinematic tree from $i$ to $j$. The absolute world position and orientation $x_{ee}(\xi;t)$ and $\phi_{ee}(\xi;t)$ of the end-effector is computed in this manner.



Any objective function involving $x_{ee}(\xi;t)$ and $\phi_{ee}(\xi;t)$ requires the Jacobians $\ppartial{x_{ee}(\xi;t)}{\xi} \in \mathbb{R}^{3 \times 6T}$ and $\ppartial{\phi_{ee}(\xi;t)}{\xi} \in \mathbb{R}^{3 \times 6T}$ for gradient-based optimization methods. Because the combination of $x_{ee}$ and $\phi_{ee}$ is a representation of the special Euclidian group \textbf{SE}(3), and $\xi_t$ is a representation of the associated lie algebra $\mathfrak{se}(3)$, deriving these Jacobians is non-trivial. Their closed form solutions are given in Appendix A.

\section{Manipulation Primitives} \label{section:primitives}
\begin{figure}
    \centering
    \includegraphics[width=0.8\columnwidth]{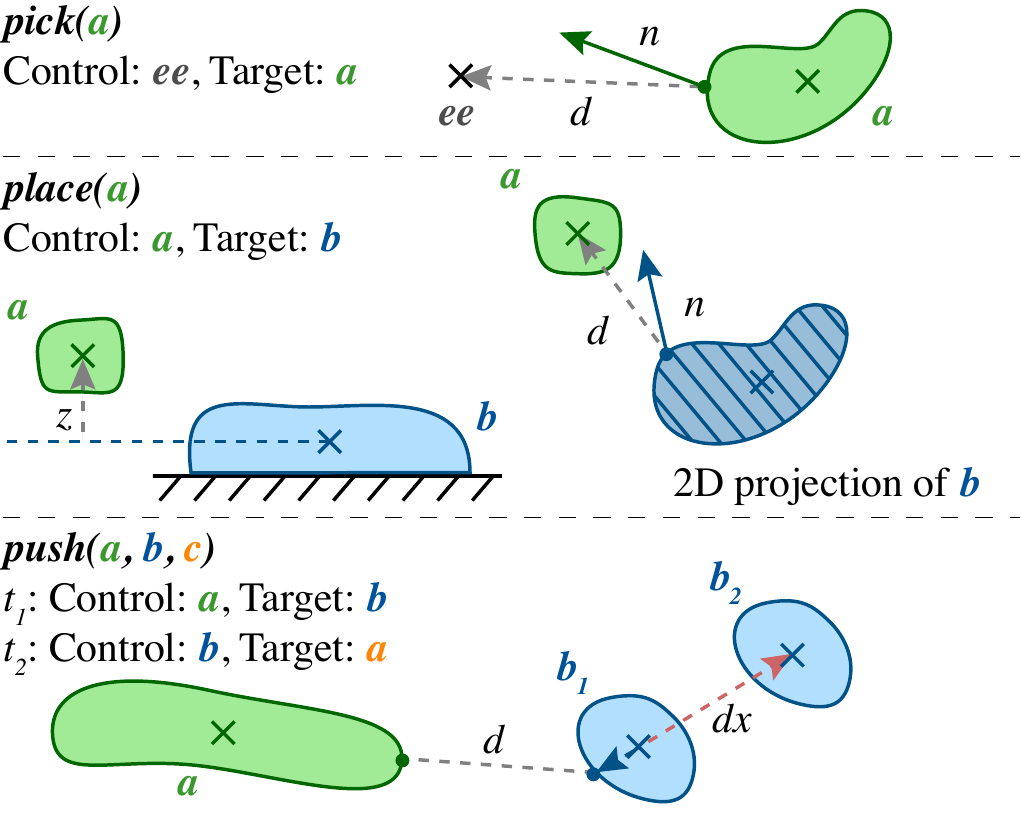}
    \caption{The $pick$ constraint requires that the signed distance between the end-effector and target is negative. The $place$ constraint requires that $a$ is above $b$ and that its center of mass is inside the support area of $b$. The $push$ constraint requires that the contact normal between $a$ and $b$ passes through the center of mass of $b$ and points in the desired direction of the push.}
    \label{fig:constraints}
\end{figure}

In this section, we define the constraint functions for each primitive manipulation action as visualized in Figure~\ref{fig:constraints}.

\subsection{Pick}

The $pick(a)$ action allows the manipulator to pick up a movable object $a$. The control frame is the end-effector and the target frame is object $a$. The preconditions are that the end-effector cannot be holding any objects before, and the postconditions are that $a$ is in the hand and no longer placed on another object. The symbolic definition is:

\textbf{Parameters}: $a$: $movable$, $obj$

\textbf{Precondition}: $\forall b. ~\neg inhand(b)$

\textbf{Postcondition}: $inhand(a) \wedge \forall b. ~\neg on(a, b)$

The constraint associated with $pick$ requires that the end-effector control point be inside the object, or the signed distance between the control point and object is negative. Here, we assume that a lower level controller will be able to perform the actual grasp given a rough desired position on the object.

Given function $\operatorname{proj}(x; a)$ that finds the closest point on the mesh of object $a$ to point $x$, and function $\operatorname{normal}(x; a)$ that finds the mesh normal of object $a$ at point $x$, we can compute the signed squared distance as follows:
\begin{subequations}
\begin{align}
    p
        &:= \operatorname{proj}(\xi_{t_p}; a) \in \mathbb{R}^3 \\
    n
        &:= \operatorname{normal}(p; a) \in \mathbb{R}^3 \\
    d
        &:= \xi_{t_p} - p \\
    f_{pick}(\xi_t; a)
        &= \frac{1}{2} \operatorname{sign}(d \cdot n) \,\|d\|_2^2 \leq 0
\end{align}
\end{subequations}

Note that $\xi_{t_p}$ represents ${}^a x_{ee}(t)$. A closed-form solution for the Jacobian of the signed squared distance can be derived by linearizing the mesh at the projection point. The Jacobian then points in the direction of the normal at the projection point.
\begin{align}
    \ppartial{f_{pick}(\xi_t; a)}{\xi_t}
        &= \begin{pmatrix} \|d\|_2 n^T & 0^{1 \times 3} \end{pmatrix}
\end{align}

\subsection{Place}

The $place(a, b)$ action allows the manipulator to place object $a$ on object $b$. The control frame is object $a$ and the target frame is object $b$. The preconditions are that $a$ must be in the end-effector and $b$ must be in the workspace of the robot. The postconditions are that $a$ is no longer in the end-effector and $a$ is on $b$. The symbolic definition is:

\textbf{Parameters}: $a$: $movable$, $obj$; $b$: $obj$

\textbf{Precondition}: $\neg eq(a, b) \wedge inhand(a) \wedge inworkspace(b)$

\textbf{Postcondition}: $\neg inhand(a) \wedge on(a, b)$

The $f_{place}$ constraint is a concatenation of $f_{touch}$, $f_{support\_area}$, and $f_{support\_normal}$, which constrains the pose of $a$ such that it is touching $b$, inside the support area of $b$, and supported by a contact normal opposite gravity.

\subsubsection{Touch}

Let $\operatorname{contact}(\xi_t; a, b)$ be a function that computes the signed distance $d$ between the closest points $p_a$ and $p_b$ between $a$ and $b$ (or the farthest points of overlap if they are interpenetrating) as output by GJK \cite{gilbert1988fast}. The constraint is:
\begin{subequations}
\begin{align}
    (d, p_a, p_b)
        &:= \operatorname{contact}(\xi_t; a, b) \in (\mathbb{R}, \mathbb{R}^3, \mathbb{R}^3) \\
    f_{touch}(\xi_t; a, b)
        &= \frac{1}{2} |d| d = 0
\end{align}
\end{subequations}

Like the $pick$ action, the linear portion of the Jacobian can be computed by linearizing the mesh at the contact point. The angular portion is approximated by finite differencing.
\begin{subequations}
\begin{align}
    n
        &:= \operatorname{normal}(p_a, a) \in \mathbb{R}^3 \\
    \ppartial{f_{touch}(\xi_t; a, b)}{\xi_{t_p}}
        &= |d| n^T
\end{align}
\end{subequations}

\subsubsection{Support Area}

To ensure that $a$ is supported by $b$, we check that the center of mass of $a$ is above the 2d projection of $b$ in the plane perpendicular to gravity. If we assume the frame of $a$ is positioned at its center of mass, then $\xi_{t_p}$ represents ${}^b x_{com_a}(t)$, the center of mass of $a$ in $b$ at timestep $t$. This constraint is then equivalent to $f_{pick}$, except in 2d.
\begin{subequations}
\begin{align}
    p
        &:= \operatorname{proj}\left(\xi_{t_p}; b\right) \in \mathbb{R}^2 \\
    n
        &:= \operatorname{normal}(p; b) \in \mathbb{R}^2 \\
    d
        &:= \xi_{t_p} - p \\
    f_{support\_area}(\xi_t; a, b)
    &= \frac{1}{2} \operatorname{sign}(d \cdot n) \|d\|_2^2 \leq 0
\end{align}
\end{subequations}

In the case where $a$ is non-convex, we split $a$ into convex regions and compute the above function with each region's center of mass. The Jacobian is:
\begin{align}
    \ppartial{f_{support\_area}(\xi_t; a, b)}{\xi_{t_p}}
        &= \begin{pmatrix} \|d\|_2 n^T & 0 \end{pmatrix}
\end{align}
If $a$ is non-convex, the angular portion of the Jacobian is computed with finite differencing. Otherwise, it is zero.

\subsubsection{Support Normal}

To ensure that the contact normal between $a$ and $b$ is pointing in the right direction (against gravity), we use a heuristic by constraining the height of $a$ to be above the center of mass of $b$. This simplification assumes that $b$ is convex and relatively flat, which means this constraint may produce a physically infeasible solution if $b$ is something like a sphere, where a surface normal above the center of mass may be pointing sideways, not against gravity. However, objects with such geometry are also difficult to use as placement targets in the real world, and can be filtered out from the optimization with symbolic preconditions.
\begin{align}
    f_{support\_normal}(\xi_t)
        &= -\xi_{t_z} \leq 0 \in \mathbb{R}
\end{align}

\subsection{Push}

The $push(a,b,c)$ action allows the manipulator to use $a$ to push $b$ on top of $c$ in cases where $b$ is outside the manipulator's workspace. The preconditions are that $a$ has to be held by the end-effector, $b$ has to be on $c$, and $c$ has to be at least partially in the workspace of the manipulator. The postcondition is that $b$ is in the manipulator's workspace, which assumes that the trajectory optimizer will be able to find a target push location inside the workspace. If such a location cannot be found, the optimizer will fail and a different action skeleton will be chosen. The symbolic definition is:

\textbf{Parameters}: $a$: $movable$, $obj$; $b$: $movable$, $obj$, $c$: $obj$

\textbf{Precondition}: $\neg eq(a, b) \wedge \neg eq(a, c) \wedge \neg eq(b, c) \wedge inhand(a) \wedge inworkspace(c)$

\textbf{Postcondition}: $inworkspace(b)$

This action is composed of two timesteps. First, the manipulator must position $a$ so that it is ready to push $b$. Second, it must push $b$ to the target position. In the first timestep, the control frame is $a$ and the target frame is $b$. In the second timestep, the control is $b$ and the target is $c$. Because $a$ is a child of $b$, it will follow $b$ wherever it is controlled.

The constraint at the first timestep is a concatenation of $f_{touch}$ from $place(a,b)$ and a $f_{push\_normal}$ constraint that constrains the vector from the point of contact to the object's center of mass to be in the direction of the push. The constraint $f_{push\_direction}$ at the second timestep restricts the target push position to be along the surface of $c$, which is assumed to be flat. Because this action involves three frames $a$, $b$, and $c$ across two timesteps, it introduces coupling between three timesteps to compute the relative poses of the three frames. The first timestep relates frame $a$ to $b$, and the second relates $b$ to $c$, but to relate $a$ to $c$, it is necessary to look at when $b$ was last placed in $c$. This coupling is limited to at most 3 timesteps, meaning the Jacobian depends on at most 18 variables.

\subsubsection{Push Normal}

Let $\operatorname{cast}(x,\delta x;a)$ be a raycasting function that finds the point of intersection between a ray with origin $x$ and direction $\delta x$ and object $b$.
\begin{subequations}
\begin{align}
    \delta x
        &:= {}^bR_c(\xi;t) \,{}^cx_b(\xi_{t+1}) - {}^bx_c(\xi;t) \in \mathbb{R}^3 \\
    p_b
        &:= {}^aR_b(\xi_t) \operatorname{cast} \left({}^b x_{com_b}, -\delta x; b \right) \in \mathbb{R}^3 \\
    p_a
        &:= \operatorname{proj}(p_b; a) \in \mathbb{R}^3 \\
    d
        &:= p_b - p_a \\
    f_{push\_normal}
        &= \frac{1}{2} \|d\|_2^2 = 0
\end{align}
\end{subequations}

The linear portion of the Jacobian at the first timestep can be simplified by assuming the point of contact $p_a$ will stay fixed to $a$ for small changes in $a$'s position. Then, the Jacobian is
\begin{align}
    \ppartial{f_{push\_normal}(\xi_t; a, b)}{\xi_{t_p}}
        &= -d^T
\end{align}

The rest of the Jacobian can be computed with finite differencing in the remaining variables.

\subsubsection{Push Direction}

If we assume that $c$ has a flat surface whose normal points in the $z$ direction, this constraint can simply constrain movement along $z$ and rotation about $x$ and $y$ to be zero. Let $s \in [-1,t)$ be the last timestep when $b$ was positioned relative to $c$, where -1 indicates that $b$ has not been manipulated yet. $\xi_s$ is the corresponding variable, or if $s = -1$, then the initial pose of $b$ in $c$.
\begin{subequations}
\begin{align}
    d\xi
        &:= \xi_{t+1} - \xi_{s} \\
    d
        &:= \begin{pmatrix} d\xi_{p_z} & d\xi_{r_x} & d\xi_{r_y} \end{pmatrix} \\
    f_{push\_direction}
        &:= \frac{1}{2} \|d\|_2^2 = 0
\end{align}
\end{subequations}

The Jacobian for the second timestep is
\begin{align}
    \ppartial{f_{push\_direction}(\xi_t; a, b)}{\xi_{t+1}}
        &= \begin{pmatrix} 0 & 0 & d^T & 0 \end{pmatrix} \label{eq:J_push_normal}
\end{align}
and the Jacobian for $s$, if it exists, is the negation of \eqref{eq:J_push_normal}.







\subsection{Collision}

Every manipulation action includes an additional constraint to ensure objects do not collide with each other. The objects can be divided into two sets: the manipulated objects $\mathcal{M}$, which consist of all objects between the end-effector and the control frame, and the environment objects $\mathcal{E}$, which consist of all other objects. Let $\operatorname{dist}$ be the signed distance between the closest points between $a$ and $b$ (equivalent to the first value in the tuple returned by $\operatorname{contact}$. The constraint function is:
\begin{align}
    f_{collision}(\xi_{1:t}; a, b)
        &= -\min_{a \in \mathcal{M},\, b \in \mathcal{E}}  \operatorname{dist}(\xi_{1:t}; a, b) \leq 0
\end{align}
The Jacobian is computed with numerical differentiation.

\section{Implementation}
\subsection{Planning}

Our STRIPS planner takes in a PDDL specification and performs a simple breadth first tree search to find candidate action skeletons. For more complex symbolic problems, a state of the art planner such as FD \cite{helmert2006fast} might be considered.

Our Cartesian frame optimizer currently leverages two different nonlinear optimizers: IPOPT \cite{biegler2009large} and NLOPT \cite{johnson2014nlopt}. Any state of the art nonlinear optimizer can be plugged in at this stage by writing wrappers for our LGP optimization API. IPOPT is an interior point method, while NLOPT features many methods. We chose the Augmented Lagrangian method with Sequential Quadratic Programming for its speed and robustness. All optimization variables are initialized to zero.

\subsection{Control}

The $pick$, $place$, and $push$ actions use operational space control \cite{khatib1987unified}. An attractive potential field in Cartesian space is created between the control frame and target frame to allow the robot to track moving targets. Repulsive potential fields are used to avoid obstacles. We choose to use torque control so that the robot can safely interact with the environment even with large perception uncertainty. Details of this controller are specified in Appendix B.

For grasping, the frame optimization simply outputs a position inside the object. To convert this position into a grasp pose for a two-fingered gripper, we use a simple heuristic that finds the top-down orientation at the optimized position with the smallest object width between the fingers. This heuristic may fail if the optimizer outputs a position that is not graspable, but it is sufficient for the simple objects in our application.

\subsection{Perception}

To track the 6-dof poses of objects, we use DBOT with particle filtering \cite{wuthrich2013probabilistic,issac2016depth} using depth data from a Kinect v2 downsampled to $320 \times 180$. Although DBOT can track objects with centimeter accuracy, even if they are partially occluded, it requires an initialization of the object poses. Furthermore, the particle filter can lose sight of objects if they move too quickly while occluded. To alleviate these two issues, we use ArUco fiducial marker tracking \cite{garrido2014automatic} as a control input to DBOT to bias the tracking particles towards the expected global pose.

\section{Results}
We present results for two different planning problems: Tower of Hanoi and Workspace Reach. The actions used in both of these problems are the ones described in Section \ref{section:primitives}. We test the framework on a 7-dof Franka Panda fitted with a Robotiq 2F-85 gripper in simulation, and then demonstrate the Workspace Reach problem in the real world. Videos of the results can be found in the supplementary material.

\begin{table}
    \centering
    \caption{IPOPT vs. NLOPT: Optimization Scores and Times}
    \label{tab:scores}
    \begin{tabular}{|l|cc|cc|}
        \hline
        \multirow{2}{*}{Solution} & \multicolumn{2}{c|}{Optimal score} & \multicolumn{2}{c|}{Time [s]} \\
        & IPOPT & NLOPT & IPOPT & NLOPT \\
        \hline
        Tower on left plate & \textbf{0.444} & --- & \textbf{3.63} & ---\\
        Tower on middle plate & \textbf{0.404} & --- & \textbf{0.96} & --- \\
        \hline
        Hook on table & \textbf{0.254} & 0.274 & 1.18 & \textbf{0.87} \\
        Hook on shelf & \textbf{0.273} & 0.358 & 3.71 & \textbf{1.42} \\
        Hook on box & 0.287 & \textbf{0.283} & 1.14 & \textbf{0.85} \\
        \hline
    \end{tabular}
\end{table}

\subsection{Tower of Hanoi}

In the Tower of Hanoi problem, there are three plates upon which towers of blocks can be formed decreasing in size from bottom to top. The task is to transfer a tower of blocks from a plate on the right side to one of two other plates: in the middle or on the left. For a tower of three blocks, the minimum number of actions needed to complete the task is 14. This problem suffers from long-term coupling between timesteps, since the placement of blocks at the bottom of the stack affects later blocks. Using relative pose variables, however, removes this coupling from the pick and place constraints.

Given a maximum symbolic tree search depth of 14, STRIPS planning finds two candidate action skeletons, each one transferring the tower to a different plate. Transferring the tower to the middle plate yields a lower optimization score than transferring to the left. This difference is also reflected in the robot's execution time for each action skeleton: transferring to the middle takes 25s while transferring to the left takes 28s.

The combined TAMP optimization takes 4.02s for this problem when using IPOPT as the nonlinear solver. As shown in Table \ref{tab:scores}, NLOPT failed to produce a feasible solution.

\subsection{Workspace Reach}

\begin{figure}
    \centering
    \includegraphics[width=\columnwidth]{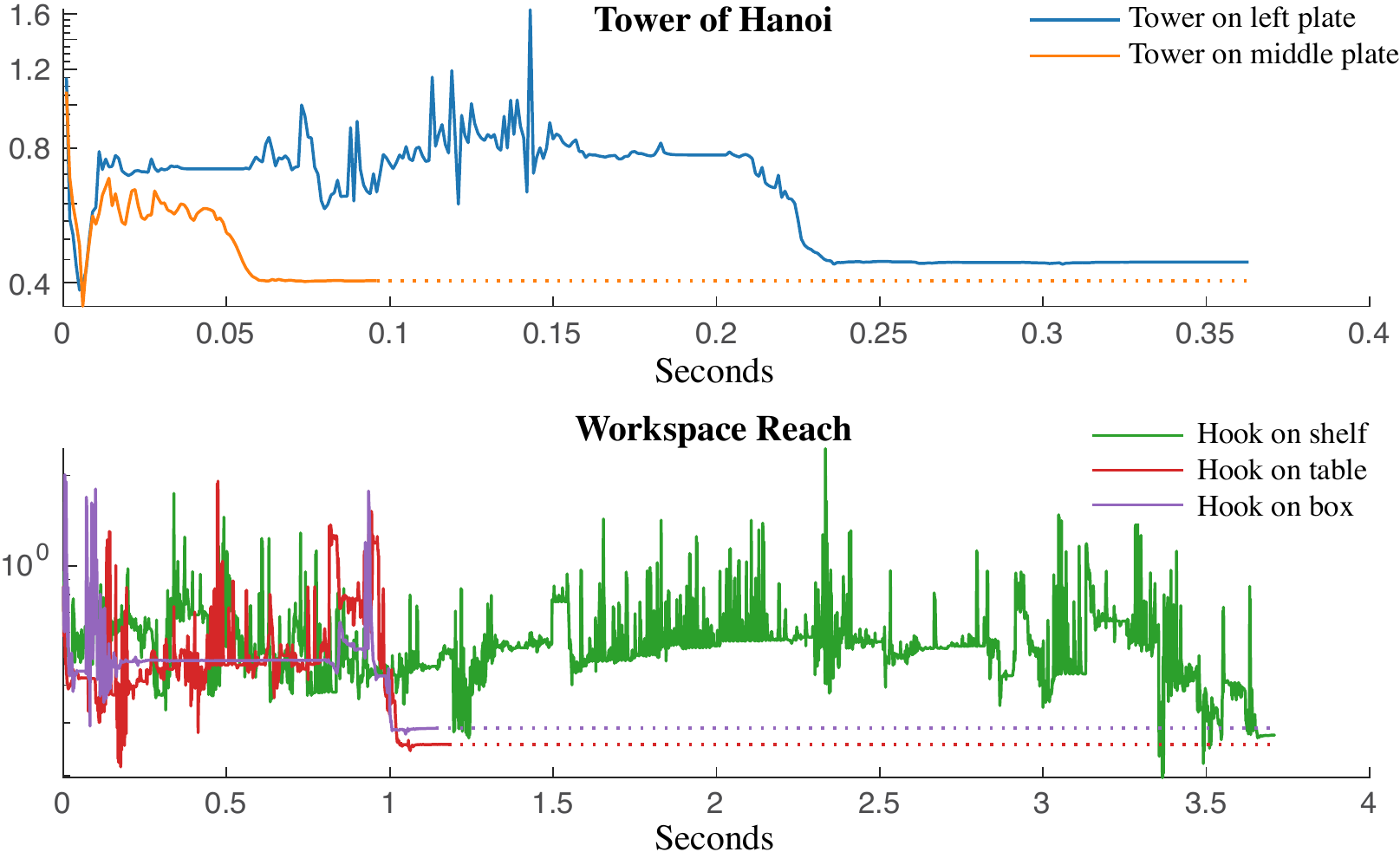}
    \caption{Optimization convergence with IPOPT. Convergence is fast, but the highly nonlinear nature of the problem makes it susceptible to local minima. The spikes occur when the IPOPT algorithm tries to escape infeasible minima.}
    \label{fig:convergence}
\end{figure}

In this problem, the robot's goal is to place a box on a shelf, but the box is outside its workspace. The world also contains a hook within the robot's workspace that can be used to bring the box closer.

With a maximum search depth of 5, there are three candidate action skeletons of the following form: 1. $pick(hook)$ 2. $push(hook, box, table)$ 3. $place(hook, \cdot)$ 4. $pick(box)$ 5. $place(box, shelf)$. The three skeletons differ in where the hook gets placed after the $push$ action: on the table, shelf, or box. Frame optimization reveals that placing the hook on the table minimizes the total distance travelled by the end-effector.

The combined TAMP optimization takes a total of 3.76s for this problem with IPOPT. Figure~\ref{fig:convergence} shows the convergence trends for the three different optimization subproblems, and Table \ref{tab:scores} compares optimization scores and times between IPOPT and NLOPT. In general, we found IPOPT to produce more optimal results with fewer constraint violations. However, it is sensitive to the initial poses of objects in the environment; it can take up to a minute to find a feasible solution, if at all, depending on the problem instance.

\section{Conclusion}
In this paper, we proposed a new TAMP algorithm based on the optimization of Cartesian frames defined relative to target objects. By design, these plans can be easily integrated with reactive closed-loop controllers that can adapt to changes in the environment in real-time and handle imperfect perception and control. While previous works on TAMP are limited to simulation or tightly controlled environments with accurate perception, we demonstrate our approach working on a real robot with noisy perception from RGB-D cameras in an environment with moving objects subject to external disturbances.

One limitation of using Cartesian frames instead of joint configurations is that robot collisions and joint limits are difficult to encode. Our TAMP method only checks object-level collisions, relying on lower level motion planners or controllers to avoid collisions with robot links and joint limits. In extreme cases, however, this heuristic would not be sufficient, and an approach that combines Cartesian and configuration variables in an augmented representation as done in \cite{lozano2014constraint} may be more appropriate.

Although this paper treats the output of the TAMP optimizer as an open-loop plan, the planning would ideally run inside a closed loop (at a frequency on the order of seconds). This way, the optimal plan would be regularly recomputed with respect to new object poses during execution. Furthermore, this would allow the framework to respond to unexpected changes in the symbolic state beyond the scope of the reactive controllers, such as objects moving in and out of the workspace or the failure of a manipulation action during execution. Creating a perception system that could detect the symbolic state of the environment is a crucial area of future research.

This paper solves LGPs by combining STRIPS planning with Nonlinear Programming, but there is potential for simplifying the optimization further for increased efficiency and robustness, such as linearizing the constraints and using a quadratic objective \cite{bemporad1999control} or decomposing the problem temporally and spatially into smaller subproblems \cite{jackson2003temporal}. 

Manipulation actions in our TAMP framework are defined by their STRIPS preconditions and postconditions, their control and target frames, and their constraint functions for optimization. The effort required to define new actions makes scalability an issue. Future work includes using learning-based approaches to alleviate the engineering bottlenecks, such as learning the preconditions and postconditions of the symbolic actions, as is considered by \citet{huang2019continuous}, or learning the constraint functions from demonstration.


%

\appendices


\section*{Acknowledgment}

Toyota Research Institute (``TRI'')  provided funds to assist the authors with their research but this article solely reflects the opinions and conclusions of its authors and not TRI or any other Toyota entity.

{\small
\bibliographystyle{IEEEtranN}
\bibliography{references}
}

\ifCLASSOPTIONcaptionsoff
  \newpage
\fi

\section{SE(3) Jacobians} \label{appendix:pose}
Objective functions used with our optimization framework will likely require the computation of object poses in world space, such as minimizing the distance travelled by the end-effector or maximizing the height of a tower of blocks. In order to use gradient-based optimization methods, it is necessary to compute the Jacobians of the object poses (in the special Euclidian group $\textbf{SE}(3)$) with respect to the optimization variables (in the Euclidian lie algebra $\mathfrak{se}(3)$). Below, we give the closed form solutions of the end-effector Jacobians $\ppartial{x_{ee}(\xi;t)}{\xi} \in \mathbb{R}^{3 \times 6T}$ and $\ppartial{\phi_{ee}(\xi;t)}{\xi} \in \mathbb{R}^{3 \times 6T}$. These Jacobians can be generalized to any object in the kinematic tree.

\subsection{Position Jacobian}
From \eqref{eq:x_xi_R_xi}, it is straightforward to derive the $3 \times 6$ block of the Jacobian $\ppartial{x_{ee}(\xi;t)}{\xi}$ corresponding to frame $i$ using the chain rule. First, let $\Lambda(i;t)$ be the set of all ancestor frames of frame $i$ at time $t$, including $i$.
\begin{align}
    \Lambda(i;t)
        &= \begin{cases}
        \{ i \} \cup \Lambda \left( \lambda(i); t \right) & \lambda(i;t) \text{ exists} \\
        \{ i \} & \text{otherwise}
        \end{cases}
\end{align}
Let $\operatorname{time}(i;t) \in \{0, \dots, t\}$ be the last timestep before $t$ where frame $i$ was manipulated, and let $\operatorname{var}(i;t)$ be the set of all last manipulated timesteps of the ancestors of $i$.
\begin{align}
    \operatorname{var}(i;t)
        &= \left\{ \operatorname{time}(j;t) ~\middle|~ j \in \Lambda(i;t) \right\}
\end{align}
Intuitively, if $s \in \operatorname{var}(i;t)$, then variable $\xi_s$ affects the global pose of frame $i$ at time $t$. Now, the $3 \times 6$ Jacobian block $\ppartial{x_{ee}(\xi;t)}{\xi_s}$ is:
\begin{align}
    \ppartial{x_{ee}}{\xi_s}
        &= \begin{cases}
            {}^0 R_{\lambda(i)}
            \begin{pmatrix}
                I^{3 \times 3} &
                \ppartial{\exp(\xi_{s_r}) \,{}^i x_{ee}}{\xi_{s_r}}
            \end{pmatrix} & s \in \operatorname{var}(i;t) \\
            0^{3 \times 6} & \text{otherwise} \\
        \end{cases}
\end{align}

The $3 \times 3$ matrix $\ppartial{\exp(\xi_{s_r}) \,{}^i x_{ee}(\xi;t)}{\xi_{s_r}}$ in the equation above is a Jacobian of the exponential map of $\mathfrak{so}(3)$. Let $\omega = \xi_{s_r} \in \mathfrak{so}(3)$ and $p = {}^i x_{ee}(\xi;t) \in \mathbb{R}^3$. This Jacobian above relates the change in the rotated vector $\exp(\omega) \,p$ to the change in $\mathfrak{so}(3)$ coordinates $\omega$. Let $\theta \in \mathbb{R}$ be the magnitude of $\omega$, or equivalently, the angle of $\omega$ as an axis-angle representation. The closed form solution of the exponential map Jacobian is given below.
\begin{align}
    \ppartial{\exp(\omega) p}{\omega}
        &= -\crossmat{\exp(\omega) \,p} d_\omega \exp(\omega) \label{eq:J_exp} \\
    d_\omega \exp(\omega)
        &= I + \frac{1 - \cos\theta}{\theta^2} \crossmat{\omega} + \frac{1 - \sin\theta}{\theta^3} \crossmat{\omega}^2
\end{align}
In the case where $\theta \approx 0$, we can use the approximation:
\begin{align}
    d_\omega \exp(\omega)
        &\approx I^{3 \times 3}
\end{align}

\subsection{Orientation Jacobian}

The angular portion of the objective function measures the angle between the orientations of two consecutive timesteps $\phi_{ee}(\xi;t-1)$ and $\phi_{ee}(\xi;t)$. Let $\Delta \phi_{ee}(\xi;t) = \phi_{ee}^{-1}(\xi; t-1) \,\phi_{ee}(\xi; t) \in \textbf{SO}(3)$. By using the formula for finding the angle of a rotation matrix, we can compute the derivative of the norm of the logarithmic map as the derivative of a matrix trace.
\begin{align}
\begin{split}
    \Delta\theta_{ee}(\xi; t)
        &= \left\| \log\left( \Delta \phi_{ee}(\xi;t) \right) \right\|_2 \\
        &= \arccos \left( \frac{1}{2} \trace{\Delta \phi_{ee}(\xi;t)} - \frac{1}{2} \right)
\end{split} \\
    \ppartial{\Delta \theta_{ee}(\xi;t)}{\xi}
        &= \frac{-1}{2 \sqrt{1 - \left( \frac{1}{2} \trace{\Delta \phi_{ee(\xi; t)}} - \frac{1}{2} \right)^2}} \ppartial{\trace{\Delta \phi_{ee}(\xi;t)}}{\xi}
\end{align}

Because the frame orientations do not depend on the position variables $\xi_{s_p}$ for $s = 1, \dots, T$, the Jacobian segments $\ppartial{\trace{\Delta \phi_{ee}(\xi;t)}}{\xi_{s_p}}$ are 0. Let $\omega = \xi_{s_r} \in \mathfrak{so}(3)$. Using the chain rule for matrices from \cite{petersen2008matrix}, we can find write the $i$th element of the gradient of the trace as:
\begin{align}
    \ppartial{\trace{\Delta \phi_{ee}(\xi;t)}}{\omega_i}
        &= \trace \left( \ppartial{\trace \Delta \phi_{ee}(\xi;t)}{\exp(\omega)}^T \ppartial{\exp(\omega)}{\omega_i} \right)
\end{align}

The rotation $\Delta \phi_{ee}(\xi;t)$ as a function of $\omega$ takes on one of the following forms, depending on the kinematic tree:
\begin{subequations}
\begin{align}
    \Delta \phi_{ee}(\xi;t)
        &= A \\
        &= B \exp(\omega) \,C \\
        &= A \exp(\omega)^{-1} \,B \\
        &= A \exp(\omega)^{-1} \,B \exp(\omega) \,C
\end{align}
\end{subequations}
where $A, B, C \in \mathbb{R}^{3 \times 3}$ are constant in terms of $\omega$. Letting $X = \exp(\omega)$, the corresponding Jacobians are:
\begin{subequations}
\begin{align}
    \ppartial{\trace \Delta \phi_{ee}(\xi;t)}{\exp(\omega)}
        &= 0^{3 \times 3} \\
        &= (BA)^T \\
        &= -\left( X^{-1} BA X^{-1} \right)^T \\
    \begin{split}
        &= \left( CA X^{-1} B - X^{-1} B X C A X^{-1} \right)^T
    \end{split}
\end{align}
\end{subequations}

Finally, the Jacobian $\ppartial{\exp(\omega)}{\omega_i}$ can be derived from \eqref{eq:J_exp} by setting $p$ to be each of the unit axis vectors.
\begin{align}
    \ppartial{\exp(\omega)}{\omega_i}
        &= \crossmat{d_\omega \exp(\omega)_i} \exp(\omega)
\end{align}
Here, $d_\omega \exp(\omega)_i$ is the $i$th column of $d_\omega \exp(\omega)$.

\newpage
\section{Operational Space Controller} \label{appendix:controllers}

The reactive behavior of operational space control at the object level makes it a natural choice for the simpler manipulation actions, like $pick$, $place$, and $push$. An attractive potential field is constructed between the control frame and target frame, and repulsive fields are constructed between the control frame and all other objects.

The attractive potential field is computed using the pose of the control and target frames obtained from perception.
\begin{subequations}
\begin{align}
    x_{des}
        &:= {}^0 T_{\operatorname{target}(t)} \exp(\xi_t) \,\,{}^{\operatorname{control}(t)} x_{ee} \\
    \ddot{x}_{goal}
        &:= -k_p (x_{ee} - x_{des}) - k_v \,\dot{x}_{ee}
\end{align}
\end{subequations}

Here, $k_p$ and $k_v$ are the position and damping gains, respectively, of the PD controller. The desired angular acceleration can be computed with analagous equations for orientation.

Because the control frame needs to be able to make contact with the target object, the repulsive field should only dampen the end-effector's velocity towards the closest obstacle.
\begin{subequations}
\begin{align}
    (d, p_a, p_b)
        &:= \min_{a \in \mathcal{M},\, b \in \mathcal{E}} \operatorname{contact}\left({}^0 x_a, {}^0 x_b; a, b \right) \\
    \hat{\delta x}_{obs}
        &:= \frac{p_b - p_a}{\|p_b - p_a\|_2} \\
    v_{obs}
        &:= \dot{x}_{ee} \cdot \hat{\delta x}_{obs} \\
    \ddot{x}_{obs}
        &:= - k_{obs} \,1[v_{obs} > 0] \,v_{obs} \,\hat{\delta x}_{obs}
\end{align}
\end{subequations}

$k_{obs}$ is a gain controlling the strength of damping towards the obstacle and $1[\cdot]$ is an indicator function that returns 1 when the condition is true and 0 otherwise.

The desired end-effector acceleration passed to the operational space controller is $\ddot{x}_{goal} + \ddot{x}_{obs}$.

\end{document}